# FrameNet CNL: a Knowledge Representation and Information Extraction Language

Guntis Barzdins

Institute of Mathematics and Computer Science, University of Latvia
Rainis blvd. 29, Riga, LV-1459, Latvia

`guntis.barzdins@lumii.lv`

**Abstract.** The paper presents a FrameNet-based information extraction and knowledge representation framework, called FrameNet-CNL. The framework is used on natural language documents and represents the extracted knowledge in a tailor-made Frame-ontology from which unambiguous FrameNet-CNL paraphrase text can be generated automatically in multiple languages. This approach brings together the fields of information extraction and CNL, because a source text can be considered belonging to FrameNet-CNL, if information extraction parser produces the correct knowledge representation as a result. We describe a state-of-the-art information extraction parser used by a national news agency and speculate that FrameNet-CNL eventually could shape the natural language subset used for writing the newswire articles.

**Keywords:** knowledge representation, information extraction, FrameNet

## 1 Introduction

In the collaborative report on the properties and prospects of Controlled Natural Languages (CNL) [7] a CNL was defined as an engineered subset of natural language such as English, which facilitates unambiguous human-human or human-machine communication. Among other uses of CNL it was stated that "CNLs appear to be particularly significant with respect to *information extraction* of and reasoning with the content of documents". As the ultimate goal of the CNL unambiguity and computability the report quoted the Leibnitz's ambition "… when there are disputes among persons, we could simply say: Let us calculate, without further ado, to see who is right".

Although mainstream effort in CNL community over past years has been devoted to defining restricted subsets of natural language, for which unambiguous translation to underlying formal representation is possible (e.g. Attempto Controlled English [4]), another research direction has focused on enhancing the CNL parsing and generation techniques to/from some Abstract Knowledge Representation (AKR) format (e.g. abstract grammar in Grammatical Framework [8]) to the point where the borderline between the natural language and CNL becomes blurred. The blurring occurs, when the *information extraction* parsers become capable of extracting the correct AKR not only from CNL, but also (to substantial degree) from the natural language (NL) documents. Meanwhile, the Grammatical Framework based text generation systems have reached the level of maturity where AKR (the result of *information extraction*) can be verbalized in the grammatically correct target language such as English.

The above overview highlights the relationship between the CNL and *information extraction* fields as illustrated in Fig. 1. In this respect the traditional formal and unambiguous CNLs can be viewed as a subset of natural language for which *information extraction* achieves 100% accuracy. The overlap between these two fields has actually been present already over several years, because FrameNet [1] – the corner-stone theory for the wide coverage *information extraction* from natural language texts, has been well represented in the CNL community already [6, 9, 11].

The purpose of this paper is to further erode the borderline between the CNL and *information extraction* approaches by defining a FrameNet CNL (FN-CNL) which actually encompasses a powerful AKR paradigm (described in Section 3) along with real-world *information extraction* system (described in Section 4). The application of FN-CNL to the real-world *information extraction* has become possible only lately (for Latvian, at least) due to the recent advances [15] in the automatic frame-semantic parsing accuracy (described in Section 2.1).

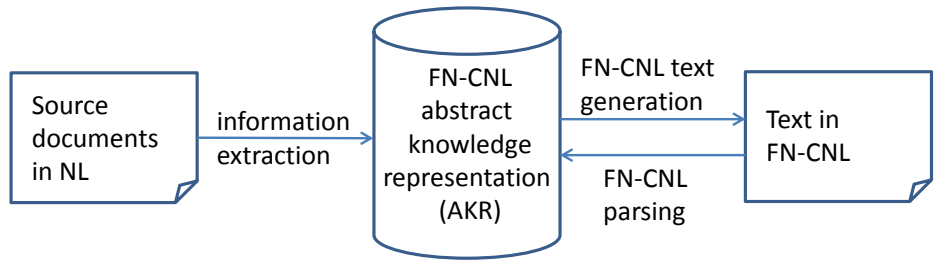

**Fig. 1.** The relationship between FN-CNL text, abstract knowledge representation (AKR) and information extraction from the natural language documents. This relationship is illustrated by a concrete example later in Fig. 7.

This paper is based on a practical *information extraction* system recently implemented for a national news agency in Latvia to extract and keep updated the biographical data profiles about publicly visible persons and organizations by automatically extracting this information from the multi-million document national newswire article archive. Although CNL was not the focus of the developed *information extraction* system, it was inspired by PAO-CNL [6], as both are based on the idea of merging FrameNet with Named Entity Linking (NEL) system to form the underlying AKR paradigm.

## 2 FrameNet

FrameNet[1] is a lexicographic database that describes word meanings based on the principles of frame semantics [1]. The central idea of frame semantics is that word meanings must be described in relation to semantic frames. Therefore, the frame and the lexical unit are the key components of FrameNet. A lexical unit is the combination of a lemma with a meaning – every new meaning of a word represents a new lexical unit. In FrameNet, each lexical unit is related to a semantic frame that it is said to evoke. The frame descriptions are coarse-grained and generalize over lexical variation. Although FrameNet addresses all parts-of-speech as frame evoking lexical units, its focus is on verbs for which the best coverage is provided.

The semantic frame describes a certain situation and the participants of that situation that are likely to be mentioned in the sentences where the evoking lexical unit (referred to as frame target) appears as illustrated by the example in Fig. 2. The semantic roles played by these participating entities are called frame elements (FE). All FrameNet frame elements are local to individual frames. This avoids the commitment to a small set of universal roles, whose specification has turned out to be controversial in the past [5]; to account for actual similarities between some frame elements (common FE such as *Time, Place*) in different frames English FrameNet includes also a rich set of frame to frame and FE to FE relations.

```
A [Durationone-year] STINT^Target [Positionas assistant lecturer]
[Employerat University College London] was followed by a
year of research in the United States.
```

**Fig. 2.** A sentence "A one-year stint as assistant lecturer at University College London was followed by a year of research in the United States" annotated with the *target* and *frame elements* of *Being employed* frame

Development of FrameNet resources for various languages is an ongoing activity [5] and in this paper Latvian and English FrameNet will be used for illustration.

### 2.1 Frame-Semantic Parsing

The benchmark methodology for frame-semantic parsing of natural language texts (sometimes regarded as automatic FrameNet Semantic Role Labelling to produce annotation as illustrated in Fig. 2) was set at SemEval-2007 [2] and specifically – by the best performing LTH system [3]. Further improvements to the methodology were implemented in the state-of-the-art SEMAFOR system [14].

To achieve high frame-semantic parsing accuracy for Latvian FrameNet (for which only a small training corpus is available) a new frame-semantic parser [15] based on the exhaustive search method nicknamed "C6.0" was developed[2]. The evaluation results for all mentioned frame-semantic parsers are shown in Table 1.

---
[1] http://framenet.icsi.berkeley.edu

**Table 1.** Evaluation results for frame target and frame element identification

|  | *Target identification* | | | *FE identification* | | |
|---|---|---|---|---|---|---|
|  | *Precision* | *Recall* | *F1* | *Precision* | *Recall* | *F1* |
| LTH (English dataset SemEval'07) | 66.2% | 50.6% | 57.3% | 51.6% | 35.4% | 42.0% |
| SEMAFOR (English dataset SemEval'07) | 69.7% | 54.9% | 61.4% | 58.1% | 38.8% | 46.5% |
| C6.0 RuleSet (English dataset SemEval'07) | **77.1%** | **53.7%** | **63.3%** | **47.3%** | **47.0%** | **47.1%** |
| C6.0 RuleSet (Latvian FrameNet) | **63.5%** | **62.7%** | **63.1%** | **65.9%** | **76.8%** | **70.9%** |

A distinct property of the C6.0 approach is that the frame-semantic parsing rules are generated in the human readable and editable format illustrated in Fig. 3 which is different from un-readable weight vectors of SVM or perceptron based machine learning algorithms used by the LTH and SEMAFOR frame-semantic parsing systems. The idea behind the exhaustive search based C6.0 algorithm was pioneered by the entropy based C4.5 and C5.0 decision-tree classification systems [16] which along with confidence limits for binomial distribution introduced also Laplace ratio $(n-m+1)/(n+2)$ for rule accuracy estimation, where $n$ is the total number of training exemplars matched by the rule and $m$ showing how many of them are false positives.

|  | Total matches | False positives | Laplace ratio |
|---|---|---|---|
| [_, _, _, _, {retaliation.n.1, punish.v.1, revengeful.s.1}, _, _, _, _, _] | 193 | 9 | 95% |
| [_, _, _, {avenger, retaliated, retaliate, avenged}, _, _, _, _, _, _] | 49 | 0 | 98% |
| [_, MD, _, get, _, _, _, _, RB, _] | 23 | 3 | 84% |
| [_, JJ, _, sanction, _, _, _, _, _, _] | 4 | 0 | 83% |
| [_, _, _, sanction, _, NNS, _, _, IN, _] | 5 | 1 | 71% |
| [_, _, #NONE#, sanction, _, _, _, ',', _, _] | 2 | 0 | 75% |

**Fig. 3.** C6.0 generated RuleSet (feature value patterns) for target word identification of frame *Revenge*. The list of features appearing in the pattern are: LEMMA, POS, NER for the previous word; LEMMA, HYPERNYM, POS, NER for the current word; LEMMA, POS, NER for the next word.

---

[2] Available at http://c60.ailab.lv

The evaluation results in Table 1 show that C6.0 based English frame-semantic parser outperforms other state-of-the-art English frame-semantic parsers, while the C6.0 based Latvian frame-semantic parser performs on par with English parsers despite smaller FrameNet annotated training corpus of 4079 sentences available for Latvian compared to 139439 sentences available for English (Latvian and English comparison is only indicative due to differences in the annotation and evaluation methodologies and the reduced number of frames in the Latvian FrameNet – see Section 2.2).

**Table 2.** Target identification F1 scores for some FrameNet frames

| Being born | 100 | Residence | 67 | Participation | 40 |
|---|---|---|---|---|---|
| Earnings and losses | 89 | Statement | 67 | Employment end | 33 |
| Death | 80 | Hiring | 62 | Product line | 33 |
| Education teaching | 71 | Membership | 50 | Lending | 29 |
| Being employed | 70 | Possession | 48 | Personal relationship | 25 |
| Change of leadership | 67 | People by vocation | 46 | Trial | 18 |
| Intentionally create | 67 | Win prize | 45 | People by origin | 16 |

The further evaluation in Table 2 breaks down the target identification accuracy for some FrameNet frames. These results illustrate that the target identification accuracy varies widely between different frame types, meaning that the low-scoring frames might convey a broader concept (which can be expressed in more ways) and thus achieving high accuracy for these frames requires a larger training corpus. Meanwhile the overall target identification accuracy above 50% still results in rather efficient information extraction from the newswire archives, because the important information tends to be duplicated multiple times in news articles (see Fig. 6) thus improving the actually perceived recall rate.

### 2.2 "Latvian" FrameNet Subset

Latvian FrameNet was created for a practical information extraction system developed for a national news agency to automatically extract biographical data about publicly visible persons and organizations mentioned in the newswire articles.

A design decision was to use a reduced number of frames – although our methodology is applicable to any number of frames, we have selected just 26 Frames out of the 1019 frames in the English FrameNet version 1.5 (*Being born, People by age, Death, Personal relationship, Being named, Residence, Education teaching, People by vocation, People by origin, Being employed, Hiring, Employment end, Membership, Change of leadership, Giving, Intentionally create, Participation, Earnings and losses, Public procurement, Possession, Lending, Trial, Attack, Win prize, Statement, Product line*) which were of interest to the national news agency; this use-case dictated also adding or removal of some frame elements (arguments) – the resulting frames are shown in Fig. 4.

Although we refer to this FrameNet subset as a "Latvian FrameNet", the information extraction approach described in this paper is equally applicable also to the English FrameNet subset of the same 26 frames.

## 3 Knowledge Representation in FN-CNL

FrameNet itself does not define any AKR paradigm – it is merely a lexicographic annotation framework. To define an AKR framework FrameNet needs to be combined with an entity identification framework, often regarded as Named Entity Linking (NEL) to create a usable AKR framework or ontology shown in Fig. 4 (this is OWLGrEd[3] visualization of the actual OWL ontology[4] used for knowledge representation in Latvian FN-CNL). Optionally, this AKR framework can further be empowered by adding an explicit time dimension as described at the end of this section.

The novelty behind the AKR framework in Fig. 4 is explicit separation of classes denoting real-world entities (light boxes) and classes denoting temporal situations captured by FrameNet frames (dark boxes). This allows AKR framework in Fig. 4 to bridge the gap between the natural language and the traditional database schemas or OWL ontologies used in information systems. From the traditional database or OWL/RDF viewpoint our AKR ontology in Fig. 4 is "non-traditional", because natural language predicates there are encoded as n-ary relations by the dark FrameNet classes, rather than by binary object-properties typical for simplistic RDF subject-predicate-object triples. As an example of n-ary predicate occurring in natural language see in Fig. 2 predicate "stint" with three arguments: duration, position, and employer.

A simplification made in the AKR framework in Fig. 4 is that only Persons and Organizations have their own dedicated light-color classes – all other frame elements are encoded by OWL data-properties of string type. This was done by purpose, because the national news agency was interested only in profiles of persons and organizations, meaning that only individuals of these classes need to be mapped to the real-world entities (which is a difficult task, discussed in Section 4). The rest of frame element fillers remain identified by the text strings as they appear in the source text.

It shall be noted that the AKR framework in Fig. 4 does not define any constraints (such as cardinality constraints – e.g., a person can have only one mother). This observation means that there is an additional conversion and constraint-checking step necessary, if the data from the AKR framework in Fig. 4 needs to be used in a more traditional database enforcing constraints on the valid data sets.

Although not yet implemented in a practical information extraction system for Latvian FrameNet, there is a further refinement possible [6] for the above described AKR framework – adding the time dimension (see Fig. 5). Note that *Time* is the dominant frame element inherited in almost all frames (see Fig. 4). For most frames extracted from the newswire texts the time of their occurrence is either explicitly specified in the text and can be retrieved by frame-semantic parser as frame element *Time* or approximate time can be retrieved from the metadata of the newswire article publication date.

---

[3] http://owlgred.lumii.lv
[4] http://www.ltn.lv/~guntis/FrameNetLV.owl

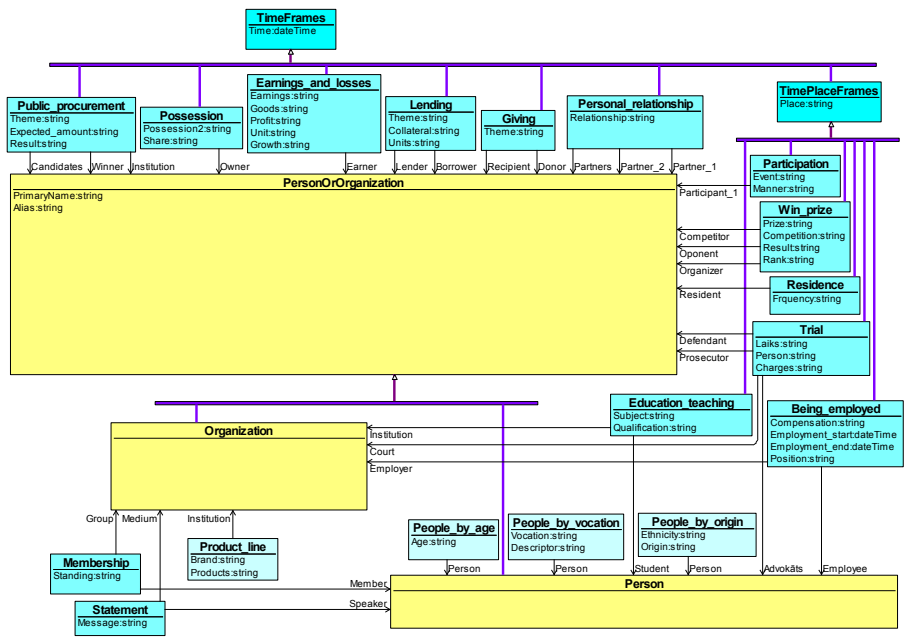

**Fig. 4.** OWLGrEd diagram of Latvian FrameNet frames (dark boxes) and Named Entity categories for frame element filler types (light boxes).

Having time associated with all extracted frames opens a possibility for avoiding seemingly contradictory facts in AKR database (e.g. *"F. Hollande is the president of France"* and *"N. Sarkozy is the president of France"*). Instead we can create a sequence of AKR database instances (e.g., one per every day of history) with each instance containing only the facts which were true on that particular day and thus make these AKR database instances internally non-contradictory (e.g. *"N. Sarkozy is the president of France"* (in DB instances for 2010) and *"F. Hollande is the president of France"* (in DB instances for 2013)). Inserting frames extracted from the text by the frame-semantic parser into the proper AKR database instance (or sequence of instances) is not an easy task [6, 10], as some frames describe an instantaneous event (e.g. frame *Attack*) while other frames describe a state which is true over prolonged period of time (e.g. frame *Being employed*). Nevertheless, resolving the time dimension (and for some sorts of tasks – also spatial movement dimension, see slides[5] from [6]) would extend the FN-CNL AKR capability to cover more of newswire text content.

---

[5] Animation on Slide 22 at http://www.semti-kamols.lv/doc_upl/polysemy.pdf

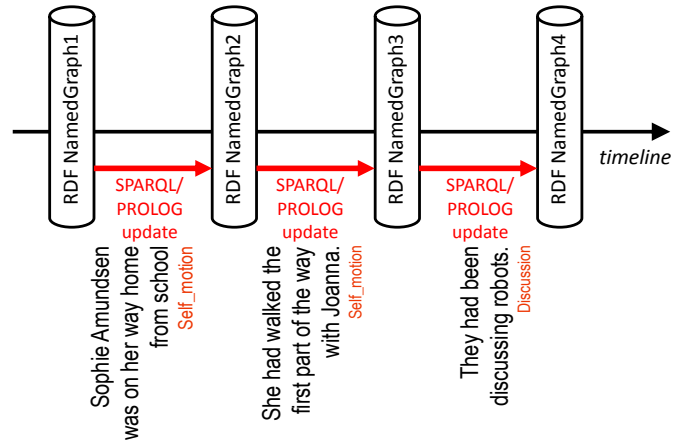

**Fig. 5.** Implementation of explicit time dimension as a sequence of AKR database copies representing the state of the world at the sequential time moments. The timeline example refers to the story illustrated in Fig. 7.

Implementation of the time dimension in the AKR framework resolves the ontology (database) versioning problem – a typical problem in simplistic ontologies or information systems loosing historic data when up-to-date information is entered (e.g. entering *"F. Hollande is the president of France"* deletes historic data *"N. Sarkozy is the president of France"*). Cross Document Coreference resolution systems (CDC, discussed in Section 4) are a good example where such historic data is useful for disambiguating entities in the documents from different time periods.

## 4 Information Extraction with FN-CNL

The increasing accuracy of frame-semantic parsing (discussed in Section 2.1) enables streamlining of information extraction task from natural language texts, such as newswire articles. Essentially the goal of such information extraction is populating the AKR ontology shown in Fig.4 with instance data retrieved from the source text. To this goal, frame-semantic parser (producing instances for the blue boxes in Fig. 4) has to be combined with Cross Document Coreference (CDC) techniques [13] to automatically determine which mentions in the text refer to the same real-world entity (instances for the yellow boxes in Fig. 4).

We have implemented such integrated information extraction system and populated it with data from approximately 1 million newswire articles. From the practical standpoint it turned out that the bottleneck of the approach is Named Entity discovery and linking accuracy – even at estimated 80% CDC accuracy it too often merged together different real-world entities with similar names or did not link together alternative spellings for the same entity, making the overall results unusable. To mitigate the problem, we deflected to the use of a predefined list of manually disambiguated well-known person and organization entities with their canonical names and common-ly used aliases, which can be identified in the text more robustly using Named Entity Linking methods similar to DBpedia Spotlight [12], but instead of DBpedia rooted in the frame instances already collected about this entity in the AKR database. Of course, this workaround links only frame elements found in the predefined list (the light class instances of ontology in Fig. 4), leaving other frame element fillers unidentified. The unidentified frame element fillers (e.g. abstractly quantified nouns or plurals) are therefore stored in simple text strings as they appear in the original sentences (technically they are stored in the same AKR database also for the light classes, only tagged as "unidentified entities").

Ieva Akuratere bija solista amatā [23]  (Ieva Akuratere had a soloist position)
Ieva Akuratere bija Puķu burves amatā [8]  (Ieva Akuratere had a Flower fairy position)
Ieva Akuratere bija mūziķes un aktrises amatā [5]  (… had a musician and actress position)
Ieva Akuratere bija deputātes amatā Rīgas domē [  (… had a member position in Riga city council)
Ieva Akuratere bija solista amatā Koncertuzvedumā [4]  (… had a soloist position in a Concert)
Ieva Akuratere bija dziedātājas amatā [3]  (… had a singer position)
Ieva Akuratere bija triju Zvaigžņu ordeņa virsnieka amatā Latvijā [3] (…had an Honor position in Latvia)

**Fig. 6.** A fragment of the automatically generated person profile (FN-CNL verbalization of *Being employed* frame). Linked Named Entities are underlined and the counts of found duplicates [in brackets] indicate the confidence level.

This mixed approach allows for creating a convenient user interface, where instance data from the AKR database in Fig. 4 is verbalized in FN-CNL using a light version of [11] producing simple FN-CNL sentences as illustrated in Fig. 6 which further can be arranged in the Curriculum Vitae like document.

## 5    FrameNet Controlled Natural Language (FN-CNL)

FN-CNL was inspired by PAO-CNL described in [6]. As illustrated in Fig. 1, FN-CNL is a verbalization of the knowledge representation database content (all or partial) by means of some FrameNet verbalization framework, such as [11].

We have implemented FN-CNL verbalization for AKR of 26 frames in Latvian FrameNet and also tested that frame-semantic parsing on this FN-CNL output achieves close to 100% accuracy (which can further be improved by hand-editing human-editable C6.0 generated frame-semantic parsing rules illustrated in Fig. 3). FN-CNL verbalization examples can potentially be used for learning unambiguous FN-CNL by human writers.

| NL text | Objects | FN Events | EN Paraphrase | LV Paraphrase |
|---|---|---|---|---|
| Sophie Amundsen was on her way home from school. | X1:Sophie Amundsen; X72:home; X73:school; X3:way; | E1:Self_motion( self_mover:X1; source:X73; goal:X72; path:X3) | E1:Sophie Amundsen moved from school to home. | E1:Sofija Amundsena pārvietojās no skolas uz mājām |
| She had walked the first part of the way with Joanna. | X4: the first part of X3; X5:Joanna; | E2: Self_motion( self_mover:X1; path:X4; co_theme:X5; time:during E1) | E2:During E1 the first part of the way Sophie Amundsen walked with Joanna. | E2: E1 laikā ceļa pirmo pusi Sofija Amundsena gāja kopā ar Jūrunu. |
| They had been discussing robots. | X6: robots; | E3: Discussion( interlocutors: X1,X5; topic:X6; time:during E2) | E3:During E2 Sophie Amundsen and Joanna discussed robots. | E3: E2 laikā Sofija Amundsena un Jūruna apsprieda robotus. |
| Joanna thought | | E4:Opinion(cognizer:X5; opinion:E5; time:during E3) | E4:During E3 Joanna stated E5. | E4: E3 laikā Jūruna apgalvoja E5. |
| the human brain was like an advanced computer. | X7:the human brain; X8: an advanced computer; | E5: Similarity( entity1:X7; entity2:X8) | E5:The human brain is similar to an advanced computer. | E5: Cilvēka smadzenes ir līdzīgas sarežģītam datoram. |

**Fig. 7.** A FN-CNL information extraction example on the left and FN-CNL verbalization examples in English and in Latvian on the right. The columns in the middle illustrate the abstract knowledge representation.

In general FN-CNL is not restricted to 26 frames of Latvian FrameNet – FN-CNL can be based on any set of frames of interest in the particular application domain thus making it adaptable to cover other linguistic or semantic domains like those currently addressed by ACE or other CNLs. Fig. 7 illustrates FN-CNL on the example of first sentences from the J.Gaardner's novel "Sophie's World" which is often used in multilingual NLP research[6]. On left is shown information extraction from the natural language resulting into AKR in the columns labeled "Object" and "FN Events". The columns on the right illustrate multilingual FN-CNL verbalization of the AKR in English and in Latvian (effectively a more formal paraphrase of the original natural language text). The paraphrase highlights the time dimension present in this example, which can be captured[7] in the knowledge representation approach illustrated in Fig. 5.

## 6 Conclusions

We have illustrated the mutually enriching relationship between the information extraction and CNL domains and described a complete natural language information extraction framework based on FN-CNL and AKR. The framework is implemented in a news agency in Latvia where it automatically extracts the profiles of public figures and organizations from newswire articles archive. As for future research we are looking into possibilities to go beyond the information extraction from the natural language texts and abstract knowledge representation (AKR) towards extracting the abstract meaning representation (AMR) [17] of the entire natural language sentences.

---
[6] http://www.language-archives.org/item/oai:tekstlab.uio.no:N10394
[7] See full example in http://attempto.ifi.uzh.ch/site/cnl2012/slides/gruzitisetal_framenet.pdf

It is interesting to note that when the information extraction frame-semantic parser is used by the national news agency, it inevitably becomes a "national parser", because the news agency uses it to evaluate the quality of articles – how high or low information extraction scores the writing of the particular journalist achieves. This stimulates editors to avoid highly ambiguous phrases in their writing and thus might be one of the first cases where a CNL starts affecting the written natural language use on the national scale.

**Acknowledgement.** This research was partially supported by the Project Nr.2DP/2.1.1.1.0/13/APIA/VIAA/014 (ERAF) "Identification of relations in newswire texts and graph visualization of the extracted relation database" under contract Nr. 1/5-2013, LU MII Nr. 3-27.3-5-2013.

## References

1. Fillmore, C.J., Johnson, C.R., Petruck, M.R.L.: Background to FrameNet. International Journal of Lexicography, 16, pp. 235—250 (2003)
2. Baker, C., Ellsworth, M., Erk, K.: SemEval-2007 task 19: Frame semantic structure extraction. In: Proceedings of SemEval-2007: 4th International Workshop on Semantic Evaluations. Prague, pp. 99–104 (2007)
3. Johansson, R., Nugues, P.: LTH: semantic structure extraction using nonprojective dependency trees. In: Proceedings of SemEval-2007: 4th International Workshop on Semantic Evaluations. Prague, pp. 227--230 (2007)
4. Fuchs N.E., Kaljurand K., Kuhn T. Attempto Controlled English for Knowledge Representation. In: Proceedings of the 4th International Reasoning Web Summer School, LNCS, vol. 5224, pp. 104–124. Springer (2008)
5. Burchardt, A. et al.: Using FrameNet for the semantic analysis of German: Annotation, representation, and automation. In: Hans C. Boas (Ed), Multilingual FrameNets in Computational Lexicography: Methods and Applications. Mouton de Gruyter, Berlin (2009)
6. Gruzitis N. and Barzdins G.: Polysemy in Controlled Natural Language Texts. In: CNL 2009 Workshop, LNCS/LNAI 5972, pp. 102–120. Springer, Heidelberg (2010)
7. Wyner A. et.al.: On Controlled Natural Languages: Properties and Prospects. In: CNL 2009 Workshop, LNCS/LNAI 5972, pp. 281-289. Springer, Heidelberg (2010)
8. Angelov K., Ranta A.: Implementing controlled languages in GF. In: CNL 2009 Workshop, LNCS/LNAI 5972, pp. 82–101. Springer, Heidelberg (2010)
9. Dannells D. Applying semantic frame theory to automate natural language template generation from ontology statements. In: Proceedings of the 6th International Natural Language Generation Conference, pp. 179-183. ACM (2010)
10. Murray, W., Singliar, T. Spatiotemporal Extensions to a Controlled Natural Language. In: CNL 2012 Workshop, LNCS/LNAI 7427, pp. 61-78. Springer, Heidelberg (2012)
11. Gruzitis N., Paikens, P., Barzdins G. 2012: FrameNet Resource Grammar Library for GF. In: CNL 2012 Workshop, LNCS/LNAI 7427, pp. 121-137. Springer, Heidelberg (2012)

12. Daiber, J., Jakob, M., Hokamp, C., Mendes, P.N.: Improving efficiency and accuracy in multilingual entity extraction, In: Proceedings of the 9th International Conference on Semantic Systems, pp. 121--124. ACM (2013)
13. Wick, M., Singh, S., Pandya, H., McCallum, A.: A Joint Model for Discovering and Linking Entities. In: Proceedings of the 2013 workshop on Automated knowledge base construction, pp. 67--72. ACM (2013)
14. Das, D., Chen, D., Martins, A.F.T, Schneider, N., Smith, N.A.: Frame-Semantic Parsing, Computational Linguistics, 40(1), pp. 9--56. (2014)
15. Barzdins, G., Gosko, D., Rituma, L., Paikens, P.: Using C5.0 and Exhaustive Search for Boosting Frame-Semantic Parsing Accuracy. In: Proceedings of the 9th Language Resources and Evaluation Conference (LREC), pp. 4476—4482. Reykjavik (2014)
16. Quinlan, J.R.: C4.5: Programs for Machine Learning. Morgan Kaufmann Publishers (1993)
17. Banarescu, L., Bonial, C., Cai, S., Georgescu, M., Griffitt, K., Hermjakob, U., Knight, K., Koehn, P., Palmer, M., and Schneider, N.: Abstract Meaning Representation for Sembanking. In: Proc. Linguistic Annotation Workshop (2013)